\documentclass[letterpaper]{article} 
\usepackage{aaai2027} 
\usepackage[hyphens]{url}  
\usepackage{graphicx} 
\usepackage{natbib}  
\usepackage{caption} 
\usepackage{algorithm}
\usepackage{algorithmic}
\usepackage{amsmath}
\usepackage{multirow}
\usepackage{graphicx}
\usepackage{subcaption}
\usepackage[table]{xcolor}
\definecolor{mygray}{RGB}{230,230,230}

\usepackage{pifont}

\usepackage{newfloat}
\usepackage{listings}
\DeclareCaptionStyle{ruled}{labelfont=normalfont,labelsep=colon,strut=off} 
\floatstyle{ruled}
\newfloat{listing}{tb}{lst}{}
\floatname{listing}{Listing}

\usepackage{booktabs}

\title{Detecting Object Hallucinations in Large Vision-Language Models via Cross-Modal Attention Drifts and Mask-Based Verification}
\author{
    Xuanbing Wen\textsuperscript{\rm 1},
    Boxu Chen\textsuperscript{\rm 1},
    Le Yang\textsuperscript{\rm 1},
    Jiakai Wang\textsuperscript{\rm 2},
    Zhengyu Zhao\textsuperscript{\rm 1},
    Chenhao Lin\textsuperscript{\rm 1},
    Chao Shen\textsuperscript{\rm 1}
}

\affiliations{
    \textsuperscript{\rm 1}Xi'an Jiaotong University, Xi'an 710049, China\\
    \textsuperscript{\rm 2}Zhongguancun Laboratory, Beijing 100194, China\\
    22009200389@stu.xidian.edu.cn,
    chenboxu@stu.xjtu.edu.cn,
    yangle15@xjtu.edu.cn,
    jk\_buaa\_scse@buaa.edu.cn,\\
    zhengyu.zhao@xjtu.edu.cn,
    linchenhao@xjtu.edu.cn,
    chaoshen@mail.xjtu.edu.cn
}

\begin{document}

\maketitle

\begin{abstract}
Despite recent advances in large vision-language models (LVLMs), object hallucination remains a major barrier to their reliable deployment. Existing detection methods often characterize visual grounding using attention from individual layers, leaving its evolution across layers underexplored. We propose CADMP, a lightweight object hallucination detection framework that combines adjacent-layer cross-modal attention drift with prediction sensitivity to targeted visual masking. During decoding, CADMP quantifies distributional changes between consecutive cross-modal attention maps to capture abrupt transitions in visual grounding. It then selects the transition with the largest drift, locates the corresponding visually relevant regions, and measures the change in prediction probability after masking these regions. These two signals provide complementary evidence: attention drift characterizes the stability of internal visual grounding, while probability variation verifies whether a prediction truly depends on the identified visual evidence. A lightweight detector integrates both signals to identify hallucinated predictions. Experiments on multiple benchmarks and representative open-source LVLMs demonstrate that CADMP achieves consistently competitive detection performance. Ablation studies further confirm the complementary contributions of adjacent-layer drift modeling and mask-based grounding verification.
\end{abstract}


\section{Introduction}

Large Vision-Language Models (LVLMs) have made substantial progress in multimodal understanding and generation~\cite{li2023llava,zhu2023minigpt}, enabling applications in areas such as autonomous driving~\cite{wei2024editable} and robotics~\cite{kim2024openvla}. Nevertheless, their reliability is still limited by object hallucination (OH), where a model mentions objects that are absent from or unsupported by the input image~\cite{bai2024hallucination}. Such errors compromise the factual consistency of model output and may pose serious
risks in safety-critical applications.

Existing hallucination assessment techniques obtain evidence from either external resources or internal model signals. External evaluation frameworks use reference annotations, auxiliary evaluators, or additional reasoning pipelines to assess image-text consistency~\cite{fu2023mme,wang2023amber,jing2024faithscore}. Although effective for offline evaluation, they may require additional annotations, models, or computational resources. Instead, internal detection methods characterize predictions using signals available within the LVLM, including predictive uncertainty~\cite{malinin2020uncertainty}, hidden representations~\cite{park2025glsim,jiang2025devils}, and cross-modal attention~\cite{zhang2025dhcp}. Among these signals, cross-modal attention is particularly relevant to visual grounding because it characterizes the interactions between output and visual tokens.

\begin{figure}[t]
  \centering
  \includegraphics[width=\linewidth]{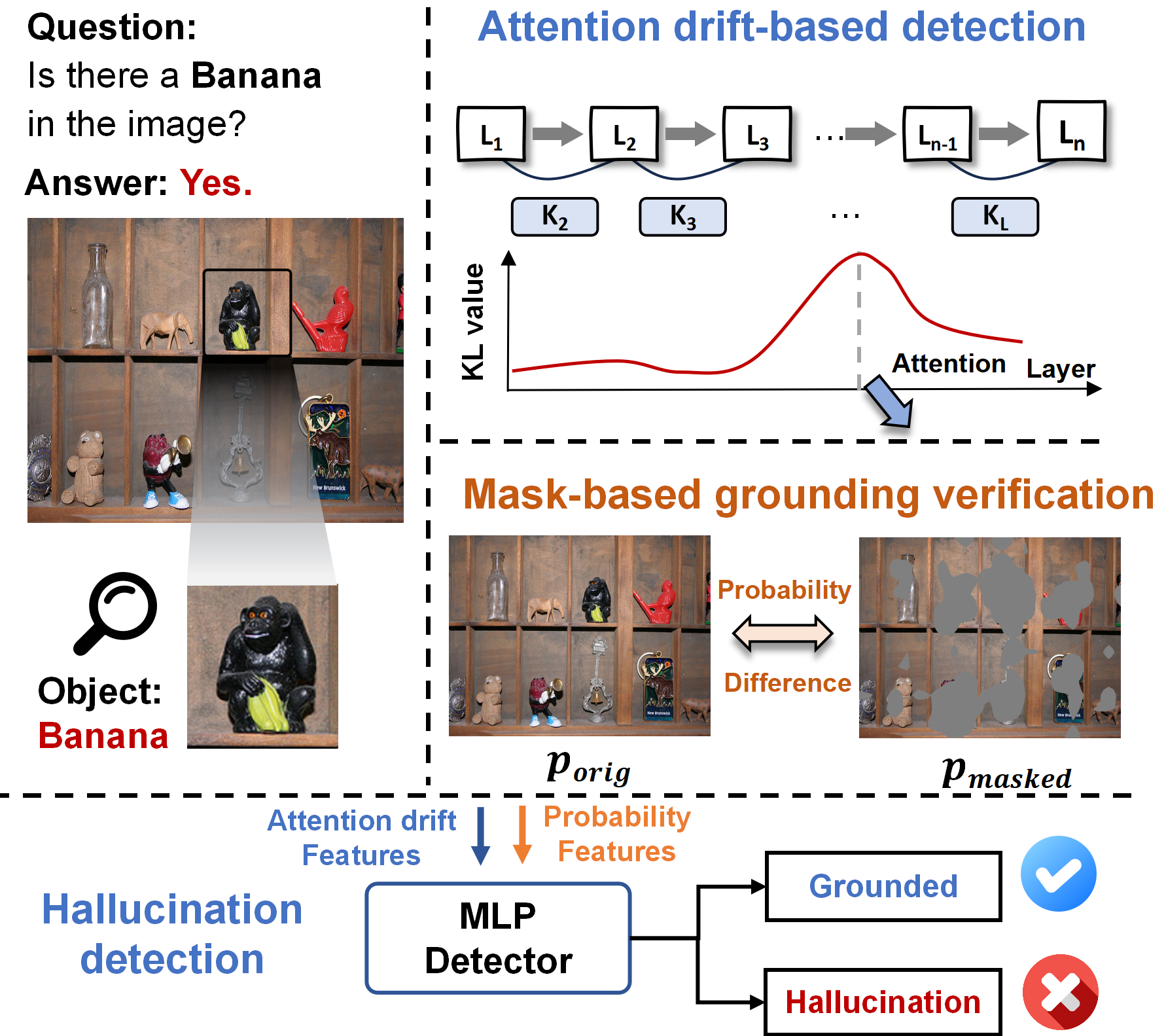}
  \vspace{-5pt}
  \caption{Overview of CADMP, detecting OHs by modeling cross-modal attention drifts and verifying visual grounding through drifted attention-guided masks.} 
  \label{fig:Object hallucination}
  \vspace{-5pt}
\end{figure}

Prior attention-based studies have used cross-modal attention patterns for hallucination detection~\cite{zhang2025dhcp}, highlighted the importance of intermediate layers in visual processing~\cite{jiang2025devils}, and modeled inter-layer changes in aggregated image-token attention for hallucination mitigation~\cite{xu2025itad}. However, changes in the full attention distribution on visual tokens between consecutive layers remain underexplored for hallucination detection. Our analysis shows that grounded and hallucinated predictions exhibit distinct layer-wise drift profiles, motivating us to treat adjacent-layer attention drift as a diagnostic signal. Based on this observation, we propose CADMP, which combines attention drift modeling with mask-based probability verification. As illustrated in Figure~\ref{fig:Object hallucination}, CADMP computes the KL divergence between consecutive cross-modal attention distributions to obtain a layer-wise drift profile. It then uses the attention map immediately before the maximum-drift transition to identify and mask relevant visual regions, and measures the resulting change in the probability of the same token. The drift profile captures the evolution of visual grounding, while the probability change estimates the prediction's dependence on visual evidence.

CADMP feeds the layer-wise drift profile, the prediction probabilities before and after masking, and their difference into a small MLP detector, while keeping the target LVLM frozen and training only the detector. Experiments on three representative open-source LVLMs across multiple hallucination benchmarks demonstrate strong and consistent detection performance. The main contributions of this work are summarized as follows:

\begin{itemize}
    \item We reveal distinct adjacent-layer attention drift patterns between grounded and hallucinated predictions, establishing attention drift as an effective detection signal.
    \item We propose CADMP, which integrates layer-wise attention drift with mask-based probability verification using a lightweight detector without retraining the LVLM.
    \item Experiments on three LVLMs and four benchmarks show that CADMP achieves the best average ACC--AUROC in eight of nine primary settings, with ablations validating both components.
\end{itemize}

\begin{figure*}[t]
  \centering
  \includegraphics[width=\linewidth]{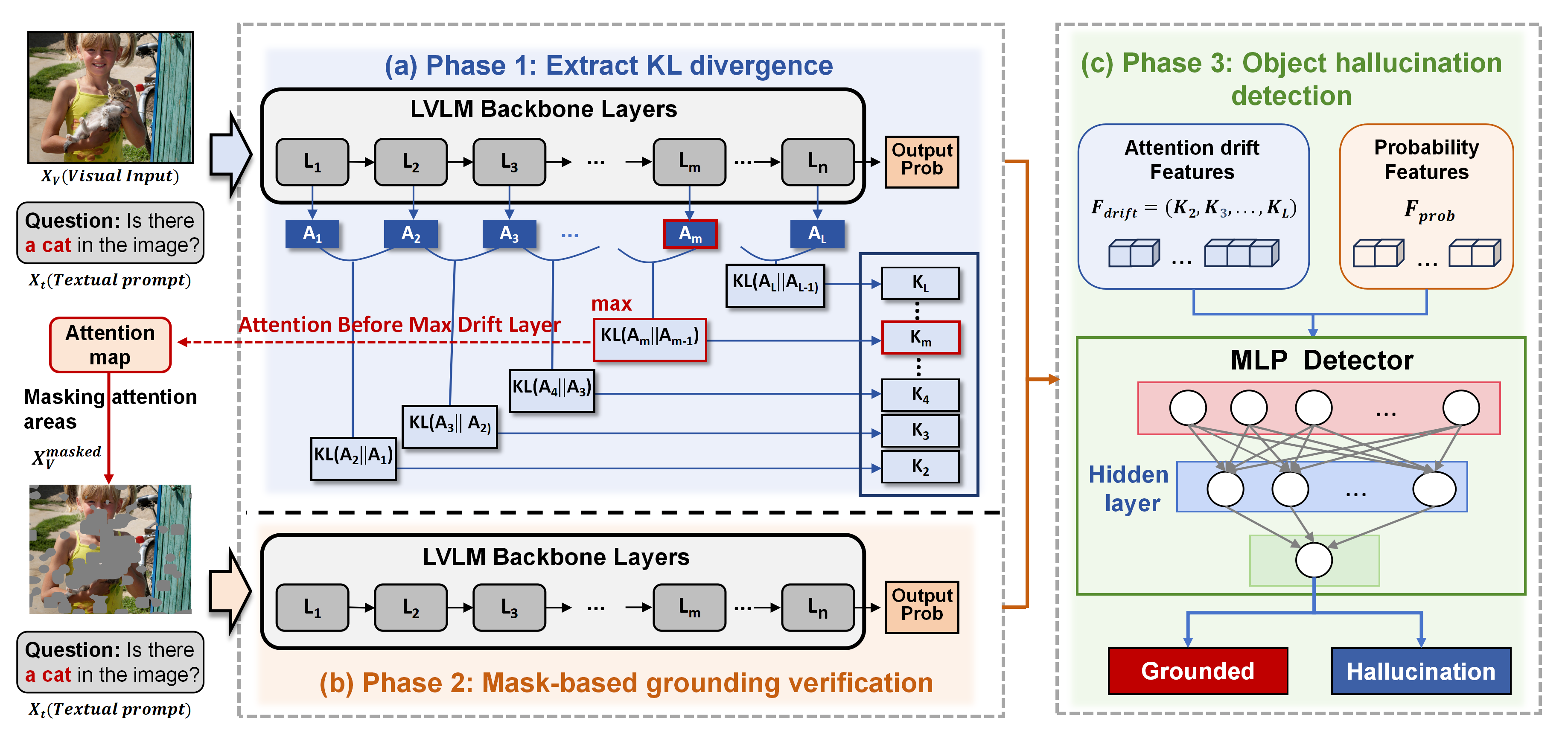}
  \vspace{-15pt}
  \caption{Overview of the proposed framework. (a) Extract KL divergence: Compute adjacent-layer KL divergences and the original target-token probability.  (b) Visual Masking: Use cross-modal attention to identify key image regions and generate masked images. (c) Hallucination Detector: Feed the extracted features into a MLP model for hallucination detection.} 
  \label{fig:overview}
  \vspace{-5pt}
\end{figure*}

\section{Related Work}

\paragraph{Large Vision-Language Models (LVLMs)} Large Vision-Language Models (LVLMs) integrate computer vision and natural language processing to enable multimodal understanding and generation~\cite{liu2023visual,bai2025qwen3,wang2024qwen2,lu2024deepseek}. An LVLM typically consists of three components: a vision encoder, a language model, and a connector. The vision encoder, such as ViT, extracts visual features from images and represents them as dense embeddings. The language model, usually a pre-trained large language model such as LLaMA~\cite{touvron2023llama}, Qwen~\cite{bai2023qwen}, or Vicuna, is responsible for language understanding and generation. The connector~\cite{li2023blip, liu2023visual, liu2024improved} bridges the visual and textual modalities by transforming visual features into representations that can be processed by the language model. Through the collaboration of these components, LVLMs achieve strong multimodal perception and reasoning capabilities.

\paragraph{Hallucination in LVLMs} Hallucination in LVLMs refers to cases where the generated output is inconsistent with  visual input or factual reality~\cite{liu2024survey}. Such errors may arise from several factors, including conflicting training data~\cite{deletang2023language}, overly strong language priors~\cite{liu2023visual}, and the loss or weakening of visual information during processing~\cite{favero2024multi}. Existing studies generally categorize hallucinations into three types: object hallucination~\cite{li2023evaluating}, attribute hallucination, and relational hallucination~\cite{jing2024faithscore}. Among them, object hallucination is the most directly related to our setting, as it occurs when the model mentions objects that do not appear in the image. By contrast, attribute hallucination involves incorrect descriptions of an object's properties, while relational hallucination concerns errors in the spatial or semantic relationships between objects.

\paragraph{Detecting Hallucination in LVLMs}
Hallucination detection in LVLMs has attracted increasing attention in recent years~\cite{wang2024mllm, augustin2025dash, arteaga2024hallucination}. Although many existing studies focus on hallucination mitigation ~\cite{hu2023ciem, you2023ferret},
accurately detecting hallucinations is equally important for improving the reliability of LVLMs. Existing detection methods mainly fall into three categories. First, some methods estimate the uncertainty of model outputs and use it as a signal for hallucination detection~\cite{zhou2023analyzing}. Second, representation-based methods such as  GLSim~\cite{park2025glsim} and IC~\cite{jiang2025interpreting} employ Logits Lens-style techniques to project hidden representations and measure their similarity to target objects for hallucination detection. Third, attention-based methods such as DHCP~\cite{zhang2025dhcp} detect hallucinations by modeling cross-modal attention patterns in internal layers. Different from these methods, our work focuses on cross-modal attention drifts during decoding and examines whether the model genuinely grounds its predictions in visual evidence, thereby enabling more precise hallucination detection.

\section{Method}
In this section, we introduce \textit{CADMP}, an object hallucination detection framework that keeps the target LVLM unchanged. As shown in Figure~\ref{fig:overview}, CADMP computes KL divergences between consecutive cross-modal attention distributions to capture layer-wise grounding drift. It then uses the attention map preceding the maximum-drift transition to mask relevant visual regions and measures the resulting probability change. Grounded predictions are expected to be more sensitive to removing visual evidence, whereas hallucinated predictions typically show smaller changes. These complementary signals are combined for hallucination detection.

\subsection{Adjacent-Layer Cross-Modal Attention Drifts}

We first investigate cross-modal attention drifts to characterize how visual grounding evolves across decoding layers. Based on previous findings that cross-modal attention patterns are important for hallucination detection~\cite{zhang2025dhcp}, we further evaluate if hallucinated predictions are more likely to exhibit abrupt changes in their attention distributions over visual tokens, and grounded predictions tend to maintain more consistent cross-layer attention patterns.

Given an image-text input pair, we focus on the cross-modal attention from the currently generated token to the visual tokens. Let $N$ denote the number of visual tokens (e.g., $N=576$ for LLaVA-1.5-7B), $H$ the number of attention heads, and $L$ the number of decoder layers (e.g., $L=32$ for LLaVA-1.5-7B). For each layer $i$, we average the cross-modal attention weights over all heads and obtain an attention vector on the visual tokens:
\[
A_i=\left[a_i^{(1)},\ldots,a_i^{(N)}\right], \  a_i^{(m)}=\frac{1}{H}\sum_{h=1}^{H}\mathrm{Attn}_{i,h}(y,v_m).
\]
$\mathrm{Attn}_{i,h}(y,v_m)$ denotes the attention weight assigned by the query position whose logits predict $y$ to the $m$-th visual token $v_m$ at layer $i$ and head $h$. By collecting the attention vectors from all layers, we obtain a layer-wise attention sequence:
\[
\mathcal{A}=\{A_1,A_2,\ldots,A_L\}.
\]
After normalization, $A_i$ can be viewed as a discrete probability distribution over the visual-token index space. Let $p_i(x)$ denote the normalized attention probability assigned to token index $x$ at layer $i$, and let $p_{i-1}(x)$ denote the corresponding distribution at layer $i-1$. To quantify the cross-layer attention drifts, we measure the discrepancy using the Kullback--Leibler (KL) divergence:
\[
K_i=D_{\mathrm{KL}}(A_i\|A_{i-1})
=\sum_{x} p_i(x)\log\frac{p_i(x)}{p_{i-1}(x)}.
\]
Here, $K_i$ measures how much the attention distribution in layer $i$ deviates from that in layer $i-1$ over the visual-token space. A larger $K_i$ indicates a stronger redistribution of attention across visual tokens between adjacent layers, while a smaller value suggests that the cross-modal attention remains relatively stable.

\begin{figure}[htp]
\vspace{-10pt}
  \centering
  \includegraphics[width=\linewidth]{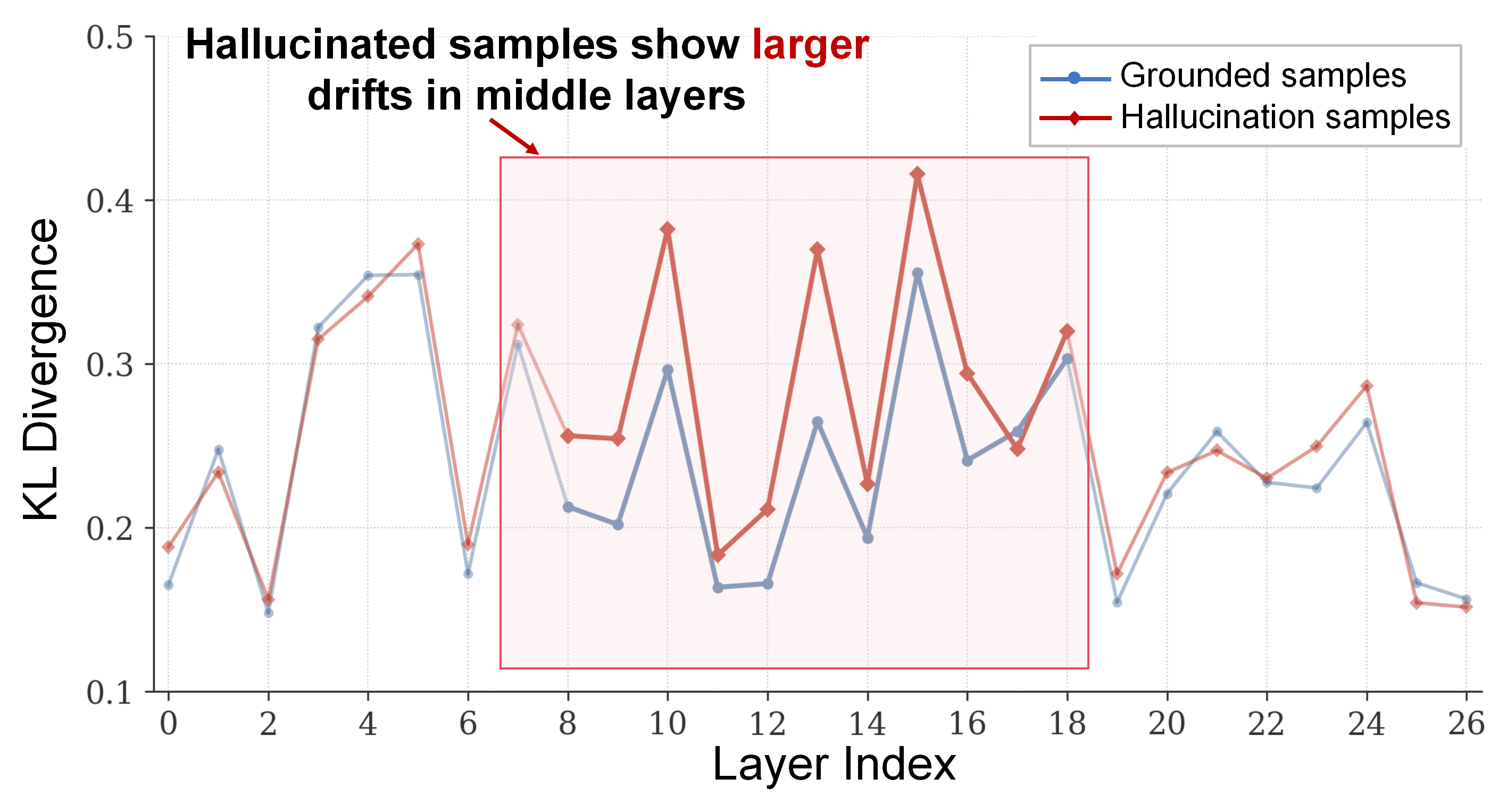}
  \vspace{-15pt}
  \caption{Average cross-modal attention drift across layers for grounded and hallucinated samples.} 
  \label{fig:Attention_Comparison}
  \vspace{-10pt}
\end{figure}

By computing the KL divergence across all consecutive layer pairs, we obtain $(L-1)$ dimensional drift features, $F_{\mathrm{drift}}=[K_2,K_3,\ldots,K_L]$. From analytical experiments in Figure~\ref{fig:Attention_Comparison}, we see whether layer-wise KL divergence provides an informative signal for hallucination detection, we analyze the drift features extracted from Qwen2.5-VL-7B on 5,000 images from MSCOCO dataset. The blue and red curves show the average attention drift for grounded and hallucinated objects. Hallucinated objects show stronger attention drift in middle layers, particularly around layers 8-16, \textit{suggesting that adjacent-layer attention drift can provide a useful signal of potential instability in visual grounding.} Therefore, We use cross-modal attention drifts to capture layer-wise variations in the model’s internal visual grounding and treat them as diagnostic signals for hallucination detection.

\subsection{Drift-guided Adaptive Masking}

The preceding analysis establishes adjacent-layer attention drift as an informative signal for hallucination detection. However, drift alone characterizes how visual attention is redistributed across layers, but does not reveal whether a prediction depends on the attended visual content. Since the drift is defined over visual-token attention distributions, the transition with the largest drift provides a principled cue for targeted visual perturbation. We use the attention map immediately before this transition, which represents the model's visual allocation before its strongest reorganization, to locate and mask highly attended regions. We then measure the probability change of the same output token. A larger decrease suggests a stronger dependence on the removed visual content, whereas a smaller change is more consistent with the reliance on language priors. Thus, drift-guided masking complements internal attention dynamics with a counterfactual measure of visual dependency.

Given an input image $v$, a text prompt $t$, a generated prefix $Y_{<t}$, and a target token $y$, we first compute its original prediction probability:
\[
P_{\mathrm{orig}} = P(y \mid v,t,Y_{<t}).
\]
Let $K_i$ denote the attention drift between layers $i-1$ and $i$, as defined in the preceding section, we select the layer before the maximum-drift transition as the masking layer:
\[
i^{*}=\arg\max_{i\in\{2,\ldots,L\}}K_i,\qquad l_{\mathrm{mask}}=i^{*}-1.
\]
The attention map at $l_{\mathrm{mask}}$ is then used to identify the visual regions most relevant to the current prediction.

After selecting $l_\mathrm{mask}$, we use its attention map to identify the visual regions most relevant to the prediction of the target token. Let $A(i,j)$ denote the attention value at spatial location $(i,j)$. We then generate a binary mask by thresholding the attention map:

\[
M(i,j)=
\begin{cases}
1, & A(i,j)\geq \mathrm{Percentile}(A,100-\tau),\\
0, & \text{otherwise},
\end{cases}
\]
In our experiments, we set $\tau=30$ , so we mask pixels at or above the 70th percentile of $A(i,j)$. We replace the selected regions with a neutral gray value of 128 while keeping the remaining regions unchanged, thereby obtaining the masked image. Detailed masking procedures are provided in the Appendix.

The masked image, together with the original prompt $t$, is then fed back into the model, and the prediction probability of the same answer token is recomputed as
\[
P_{\mathrm{masked}} = P(y \mid v_{\mathrm{masked}}, t, Y_{<t}).
\]
Importantly, we compare the probability of the same token under the original and masked visual conditions, which provides a direct measure of confidence change after removing the relevant visual evidence. Finally, we define the probability variation as
\[
\Delta P = P_{\mathrm{orig}} - P_{\mathrm{masked}},
\]


\subsection{Hallucination Detection}
Leveraging the differences in cross-modal attention drifts and probability variations between grounded and hallucinated samples,  we design a detector to verify the factuality of the answer produced by the model. We train a lightweight MLP detector by feeding it the extracted KL-divergence profile together with the prediction probabilities before and after masking. For example, LLaVA-1.5-7B contains 32 layers, producing 31 adjacent-layer KL-divergence values. We then put the 31-dimensional feature vector, together with $P_{\mathrm{orig}}$, $P_{\mathrm{masked}}$ and $\Delta P$ , into the MLP detector. The detector outputs a hallucination score for each object prediction, which is converted into a probability through a sigmoid function. 

To address the class imbalance between hallucinated and grounded samples, we employ weighted random sampling during training, assigning each sample a weight inversely proportional to the size of its class.

\section{Experiments}

\subsubsection{Datasets}
To evaluate the effectiveness and generalizability of the proposed method for hallucination detection in LVLMs, we conduct experiments on four datasets. Specifically, we use POPE~\cite{li2023evaluating} and Pascal VOC~\cite{everingham2010pascal} to evaluate object-existence hallucination detection.
In addition, we construct a COCO-Caption setting based on MSCOCO~\cite{lin2014microsoft}, where LVLM-generated captions are annotated using the CHAIR metric~\cite{rohrbach2018object} to evaluate descriptive object hallucination detection. 
We also conduct experiments on AMBER~\cite{wang2023amber}, which covers object, attribute, and relation hallucinations. More details about the datasets are provided in the supplementary materials.

\begin{table*}[t]
  \centering
  \fontsize{10}{12}\selectfont
  \setlength{\tabcolsep}{6pt}
  \renewcommand{\arraystretch}{1.05}

  \begin{tabular}{llcc|cc|cc}
    \toprule
    \multirow{2}{*}{Model} &
    \multirow{2}{*}{Method} &
    \multicolumn{2}{c|}{COCO-Cap.} &
    \multicolumn{2}{c|}{POPE} &
    \multicolumn{2}{c}{VOC} \\
    \cmidrule(lr){3-4}
    \cmidrule(lr){5-6}
    \cmidrule(lr){7-8}
    & & ACC & AUROC
      & ACC & AUROC
      & ACC & AUROC \\
    \midrule

    \multirow{7}{*}{LLaVA-1.5-7B}
    & IC (ICLR'25) & $77.7_{\pm0.5}$ & $76.5_{\pm0.6}$ & $73.6_{\pm0.4}$ & $73.7_{\pm0.3}$ & $80.2_{\pm0.3}$ & $79.9_{\pm0.6}$ \\
    & GLSim (NeurIPS'25) & $74.7_{\pm0.4}$ & $52.9_{\pm1.5}$ & $71.8_{\pm0.7}$ & $62.7_{\pm0.5}$ & $78.7_{\pm0.6}$ & $63.4_{\pm0.7}$ \\
    & NLL (ICLR'24) & $76.0_{\pm0.5}$ & $71.5_{\pm0.4}$ & $78.2_{\pm0.1}$ & $79.2_{\pm0.1}$ & $90.6_{\pm0.3}$ & $76.7_{\pm0.2}$ \\
    & Entropy (ICLR'21) & $75.3_{\pm0.1}$ & $72.6_{\pm0.4}$ & $80.6_{\pm0.5}$ & $81.1_{\pm0.4}$ & $92.2_{\pm0.1}$ & $85.8_{\pm0.1}$ \\
    & SVAR (CVPR'25) & $75.4_{\pm1.0}$ & $80.9_{\pm0.7}$ & $62.5_{\pm0.7}$ & $70.2_{\pm0.8}$ & $72.2_{\pm0.6}$ & $85.9_{\pm0.5}$ \\
    & DHCP (MM'25) & $76.4_{\pm0.7}$ & $81.8_{\pm0.6}$ & $70.6_{\pm0.2}$ & $60.7_{\pm1.2}$ & $91.0_{\pm0.2}$ & $92.2_{\pm0.3}$ \\
    & \cellcolor{mygray}\textbf{CADMP (Ours)}
      & \cellcolor{mygray}\textbf{\boldmath$82.0_{\pm0.2}$}
      & \cellcolor{mygray}\textbf{\boldmath$87.4_{\pm0.2}$}
      & \cellcolor{mygray}\textbf{\boldmath$84.9_{\pm1.1}$}
      & \cellcolor{mygray}\textbf{\boldmath$81.4_{\pm0.2}$}
      & \cellcolor{mygray}\textbf{\boldmath$94.5_{\pm0.2}$}
      & \cellcolor{mygray}\textbf{\boldmath$94.9_{\pm0.2}$} \\
    \midrule

    \multirow{7}{*}{Qwen2.5-VL-7B}
    & IC (ICLR'25) & $86.7_{\pm0.5}$ & $66.1_{\pm2.9}$ & $55.3_{\pm1.5}$ & $53.4_{\pm1.4}$ & $75.4_{\pm0.4}$ & $72.2_{\pm0.5}$ \\
    & GLSim (NeurIPS'25) & $86.7_{\pm1.3}$ & $68.3_{\pm0.8}$ & $60.4_{\pm0.6}$ & $63.8_{\pm0.5}$ & $70.8_{\pm0.3}$ & $76.8_{\pm0.7}$ \\
    & NLL (ICLR'24) & $86.9_{\pm0.7}$ & $63.6_{\pm2.7}$ & $84.2_{\pm0.4}$ & $80.6_{\pm0.2}$ & $88.6_{\pm0.8}$ & $86.9_{\pm0.1}$ \\
    & Entropy (ICLR'21) & $86.7_{\pm0.8}$ & $65.4_{\pm3.4}$ & $84.6_{\pm0.5}$ & $80.8_{\pm0.3}$ & $89.9_{\pm0.4}$ & $90.9_{\pm0.5}$ \\
    & SVAR (CVPR'25) & $86.9_{\pm0.5}$ & $74.2_{\pm0.4}$ & $84.5_{\pm0.1}$ & $70.1_{\pm0.4}$ & $87.1_{\pm0.5}$ & $67.0_{\pm0.6}$ \\
    & DHCP (MM'25) & $86.6_{\pm0.6}$ & $72.0_{\pm2.2}$ & \textbf{\boldmath$86.8_{\pm0.3}$} & $74.7_{\pm0.6}$ & $94.0_{\pm0.1}$ & $91.2_{\pm0.4}$ \\
    & \cellcolor{mygray}\textbf{CADMP (Ours)}
      & \cellcolor{mygray}\textbf{\boldmath$87.6_{\pm0.7}$}
      & \cellcolor{mygray}\textbf{\boldmath$82.6_{\pm1.0}$}
      & \cellcolor{mygray}$85.8_{\pm0.8}$
      & \cellcolor{mygray}\textbf{\boldmath$82.5_{\pm0.5}$}
      & \cellcolor{mygray}\textbf{\boldmath$96.5_{\pm0.1}$}
      & \cellcolor{mygray}\textbf{\boldmath$92.1_{\pm0.6}$} \\
    \midrule

    \multirow{7}{*}{InternVL-2.5-4B}
    & IC (ICLR'25) & $87.1_{\pm0.3}$ & $72.7_{\pm0.9}$ & $78.5_{\pm0.3}$ & $61.6_{\pm0.3}$ & $77.7_{\pm0.8}$ & $71.3_{\pm0.8}$ \\
    & GLSim (NeurIPS'25) & $87.2_{\pm0.3}$ & $61.5_{\pm2.5}$ & $70.7_{\pm0.8}$ & $60.6_{\pm1.1}$ & $72.6_{\pm0.6}$ & $61.1_{\pm0.7}$ \\
    & NLL (ICLR'24) & $87.3_{\pm0.4}$ & $61.8_{\pm1.4}$ & $86.0_{\pm0.1}$ & $82.5_{\pm0.2}$ & $95.1_{\pm0.4}$ & $89.0_{\pm0.1}$ \\
    & Entropy (ICLR'21) & $87.3_{\pm0.3}$ & $62.4_{\pm1.3}$ & $86.9_{\pm0.7}$ & \textbf{\boldmath$83.2_{\pm0.4}$} & $92.8_{\pm0.1}$ & $89.9_{\pm0.9}$ \\
    & SVAR (CVPR'25) & $87.4_{\pm0.5}$ & $82.5_{\pm1.8}$ & $70.1_{\pm1.0}$ & $62.5_{\pm0.8}$ & $80.6_{\pm0.1}$ & $72.2_{\pm1.4}$ \\
    & DHCP (MM'25) & $87.2_{\pm0.3}$ & $78.8_{\pm0.9}$ & $72.5_{\pm0.5}$ & $72.2_{\pm0.5}$ & $67.6_{\pm0.2}$ & $78.8_{\pm1.2}$ \\
    & \cellcolor{mygray}\textbf{CADMP (Ours)}
      & \cellcolor{mygray}\textbf{\boldmath$89.0_{\pm0.4}$}
      & \cellcolor{mygray}\textbf{\boldmath$87.1_{\pm1.1}$}
      & \cellcolor{mygray}\textbf{\boldmath$88.2_{\pm1.1}$}
      & \cellcolor{mygray}$81.1_{\pm1.5}$
      & \cellcolor{mygray}\textbf{\boldmath$97.5_{\pm0.1}$}
      & \cellcolor{mygray}\textbf{\boldmath$92.8_{\pm0.4}$} \\
    \bottomrule
  \end{tabular}
 \vspace{-5pt}
  \caption{Comparison of object hallucination detection performance in terms of accuracy (ACC) and area under the ROC curve (AUROC) across three models and three datasets. Higher values indicate better performance.}
  \label{tab:acc_AUROC_combined}
  \vspace{-5pt}
\end{table*}

\subsubsection{Evaluation Metrics}
To evaluate detection performance, we report two metrics: (1) the area under the receiver operating characteristic curve (AUROC), (2) classification accuracy (ACC). 
These metrics are commonly used in OH detection and provide complementary views of model performance~\cite{park2025glsim,hoangxuan2026pas}.

\noindent \textbf{Implementation.} We evaluate our method on three representative LVLMs: LLaVA-1.5-7B~\cite{liu2023visual,liu2024improved}, Qwen2.5-VL-7B~\cite{bai2025qwen25vl}, and InternVL-2.5-4B~\cite{chen2024internvl}. Also, we compare the proposed CADMP with recent OH detection baselines, including Internal Confidence (IC)~\cite{jiang2025interpreting}, Global-Local Similarity (GLSim)~\cite{park2025glsim}, Negative Log-likelihood (NLL)~\cite{zhou2023analyzing}, a token probability-based uncertainty method denoted as Entropy~\cite{malinin2020uncertainty}, summed Visual Attention Ratio (SVAR)~\cite{jiang2025devils}, and the cross-modal attention pattern-based method DHCP~\cite{zhang2025dhcp}. For fair comparison, all methods are evaluated on the same training and test splits under the same experimental protocol. More details are provided in the supplementary materials.


\subsection{Experimental results}

\subsubsection{Results on OH Detection}



As shown in Table~\ref{tab:acc_AUROC_combined}, CADMP achieves the best result in 16 of the 18 model--dataset--metric combinations, demonstrating consistent effectiveness across architectures and evaluation settings. The gains are particularly pronounced on COCO-Caption, where CADMP improves AUROC over the strongest baseline by 5.6, 8.4, and 4.6 percentage points for LLaVA-1.5-7B, Qwen2.5-VL-7B, and InternVL-2.5-4B, respectively. On Pascal VOC, CADMP ranks first in both ACC and AUROC across all three models, reaching up to 97.5\% ACC and 94.9\% AUROC. It also remains competitive on POPE, although its ACC on Qwen2.5-VL-7B and AUROC on InternVL-2.5-4B are slightly below the corresponding best baselines. Overall, these results indicate that combining adjacent-layer attention drift with mask-based probability verification provides robust and complementary signals for object hallucination detection.

\begin{figure*}[h]
  \centering
  \includegraphics[width=\linewidth]{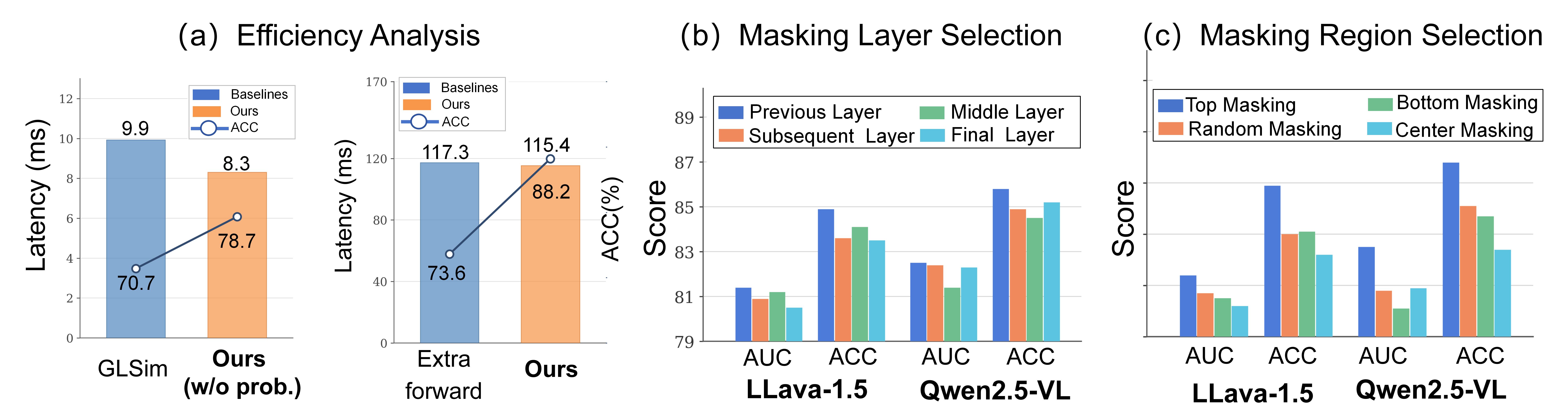}
  \vspace{-5pt}
  \caption{(a) Efficiency analysis on InternVL-2.5-4B using the POPE dataset; (b) Different masking-layer selection strategies; (c) Different masking-region selection strategies.} 
  \label{fig:efficiency}
  \vspace{-5pt}
\end{figure*}

\subsubsection{Results on AMBER} 

To further evaluate whether CADMP generalizes beyond OH detection, we conducted experiments on the AMBER discriminative dataset, which covers three types of hallucinations: object, attribute, and relation hallucinations. As shown in Table~\ref{tab:amber}, CADMP consistently outperforms representative baselines across all three hallucination types on Qwen2.5-VL-7B. These results suggest that the proposed attention-drift modeling and adaptive mask-based grounding verification are not limited to object hallucination, but can also provide useful detection signals for more fine-grained hallucination types.

\begin{table}
\centering
\small
\setlength{\tabcolsep}{4pt}
\begin{tabular}{l p{0.9cm}<{\centering} p{0.9cm}<{\centering}p{0.9cm}<{\centering}p{0.9cm}<{\centering}p{0.9cm}<{\centering}p{0.9cm}<{\centering}}
\toprule
\multirow{2}{*}{Method}
& \multicolumn{2}{c}{Object}
& \multicolumn{2}{c}{Attribute}
& \multicolumn{2}{c}{Relation} \\
\cmidrule(lr){2-3}
\cmidrule(lr){4-5}
\cmidrule(lr){6-7}
& ACC & AUROC
& ACC & AUROC
& ACC & AUROC \\
\midrule
Entropy& 92.89 &86.92 &82.16  &78.99 &58.59  &55.38  \\
DHCP& 93.51 & 81.67& 74.19 & 70.96 & 53.91 & 54.89 \\
GLSim& 87.54 & 86.76& 71.35 & 67.52& 56.25 & 60.19 \\
\cellcolor{mygray} \textbf{CADMP}& \cellcolor{mygray}\textbf{93.69} & \cellcolor{mygray}\textbf{87.11}& \cellcolor{mygray}\textbf{82.50} & \cellcolor{mygray}\textbf{80.75}& \cellcolor{mygray}\textbf{72.40} & \cellcolor{mygray}\textbf{73.05} \\
\bottomrule
\end{tabular}
 \vspace{-5pt}
\caption{Hallucination detection performance on the AMBER benchmark using Qwen2.5-VL-7B.}
\label{tab:amber}
 \vspace{-10pt}
\end{table}

\subsubsection{Efficiency Analysis} To evaluate the practical overhead of CADMP, we performed an efficiency analysis covering latency and ACC (Figure~\ref{fig:efficiency}). Compared to GLSim~\cite{park2025glsim}, our method without additional masked-image forward pass can achieve higher accuracy with lower latency. Moreover, compared to the forward based detection method in~\cite{liu2023mitigating}, our method still achieves an accuracy of 88.2\% with faster inference speed, demonstrating the effectiveness of CADMP.

\begin{table}[htp]
\centering
\small
\begin{tabular}{l l p{1.1cm}<{\centering} p{1.1cm}<{\centering}}
\toprule
\textbf{Model} & \textbf{Setting} & \textbf{DHCP} & \textbf{CADMP} \\
\midrule
\multirow{2}{*}{LLaVA-1.5-7B}
& POPE $\rightarrow$ VOC & 77.8 & \textbf{86.7} \\
& VOC $\rightarrow$ POPE & 65.9 & \textbf{74.8} \\
\midrule
\multirow{2}{*}{InternVL-2.5-4B}
& POPE $\rightarrow$ VOC & 78.9 & \textbf{82.6} \\
& VOC $\rightarrow$ POPE & 71.8 & \textbf{75.5} \\
\midrule
\multirow{5}{*}{Qwen2.5-VL-7B}
& VOC $\rightarrow$ POPE & 71.2 & \textbf{74.4} \\
& POPE $\rightarrow$ VOC & 73.1 & \textbf{87.6} \\
& POPE $\rightarrow$ Am-Obj & 70.2 & \textbf{73.2} \\
& POPE $\rightarrow$ Am-Attr & 72.5 & \textbf{76.7} \\
& POPE $\rightarrow$ Am-Rel & 50.4 & \textbf{66.4} \\
\bottomrule
\end{tabular}
\caption{Cross-dataset AUROC results comparison between DHCP and our method across three LVLMs.}
\label{tab:cross_dataset_compare}
\vspace{-10pt}
\end{table}

\subsubsection{Cross-dataset Generalization Analysis.} 
To further evaluate whether CADMP captures intrinsic hallucination cues rather than dataset-specific patterns, we train the detector on a source dataset and then evaluate it on different target datasets without any additional adaptation (Source $\rightarrow$ Target). As shown in Table~\ref{tab:cross_dataset_compare}, our method consistently outperforms DHCP across all target datasets and all hallucination types. For example, CADMP obtains AUROC scores of 73.2\% on AMBER-object,  76.7\% on AMBER-Attribute and 66.4\% on AMBER-relation, demonstrating its ability to generalize across diverse types of hallucinations. It also maintains favorable performance on Pascal VOC. These results suggest that CADMP captures more robust and transferable hallucination signals, rather than relying primarily on dataset-specific characteristics.







\subsection{Ablation Studies}

We further conducted a series of ablation studies for the proposed CADMP detection method to evaluate the contribution of each component.

\begin{table}[htbp]
\centering
\small
\begin{tabular}{lp{1.2cm}<{\centering}p{1.4cm}<{\centering}p{1.4cm}<{\centering}p{1.0cm}<{\centering}}
\toprule
\textbf{Metric} & {\scriptsize \textbf{LLaVA-1.5}} & {\scriptsize\textbf{Qwen2.5-VL}} & {\scriptsize\textbf{InternVL-4B}} & Avg. \\ 
\midrule
Cosine          & 86.6   & 80.8    &86.5   &  84.6     \\
Euclidean       & 86.4   & 80.6    &86.9    &   84.6          \\
Pearson         & 86.7   &  80.5   &86.7     & 84.6       \\
KL   & \textbf{87.5} & \textbf{81.5} & \textbf{87.9} & \textbf{85.6}\\
\bottomrule
\end{tabular}
\caption{Comparison of AUROC across different cross-modal attention drift methods on the MSCOCO dataset.}
\vspace{-10pt}
\label{tab:hallucination_detection}
\end{table}

\subsubsection{Selection of cross-layer evaluation metrics} As shown in Table~\ref{tab:hallucination_detection}, we compare several methods for measuring cross-modal attention drift, including Cosine Similarity, Euclidean Distance, Pearson Correlation, and Kullback--Leibler (KL) Divergence, across different models on the MSCOCO dataset. Among them, KL divergence achieves the best overall performance, with an average AUROC of 85.6\%, outperforming the other methods. This result is consistent with our formulation of cross-modal attention drift as redistribution over visual tokens and suggests that KL divergence is better suited to capture abnormal grounding transitions.


\subsubsection{Layer Selection for Cross-modal Attention Masking} 
To validate whether the layer immediately preceding the largest attention drift can better capture useful grounded information, we conduct an ablation study using four methods: the previous layer of the largest drift appearance, the subsequent layer, a fixed middle layer and the final layer. As shown in Figure~\ref{fig:efficiency} (b), experiments on LLaVA-1.5 and Qwen2.5-VL show that selecting the previous layer achieves the best performance. These results suggest that the attention distribution immediately before the most abrupt cross-layer redistribution may provide a somewhat more informative signal for grounding verification.






\begin{figure*}[h]
  \centering
  \includegraphics[width=\linewidth]{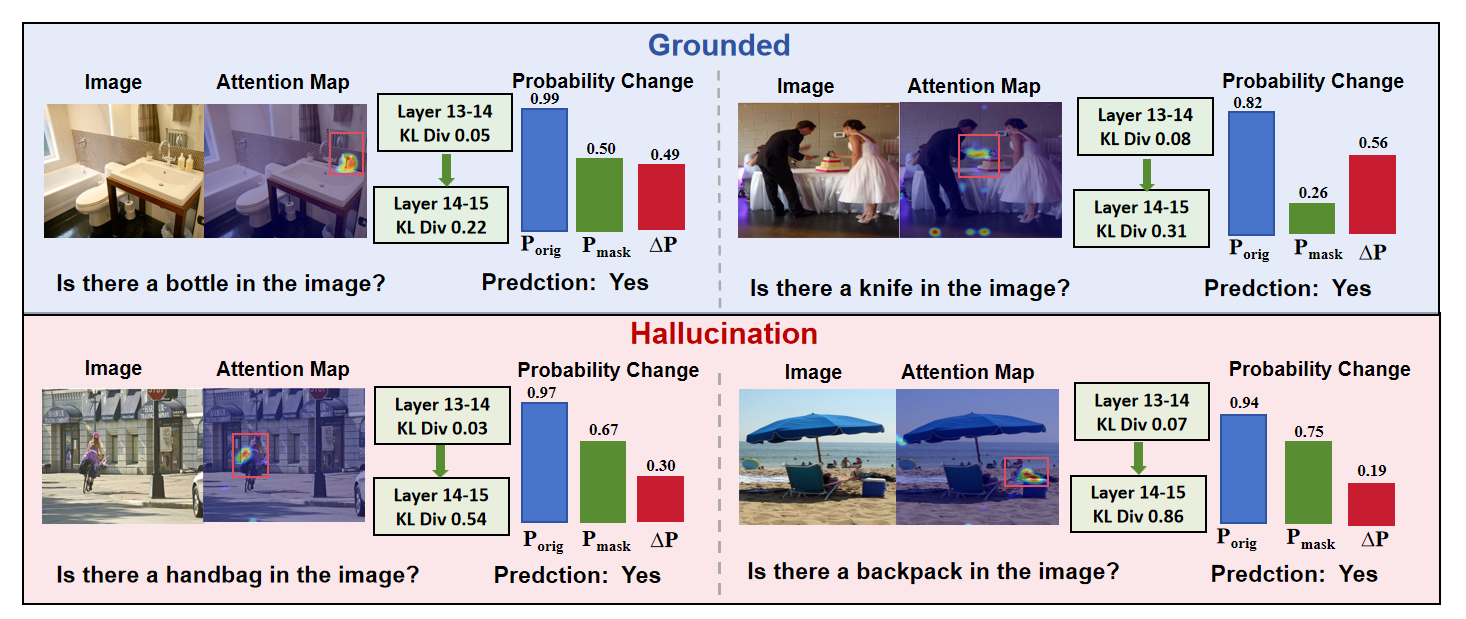}
  \caption{CADMP visualization of grounded (top) and hallucination (bottom) cases.} 
  \label{fig:case_analysis}
  \vspace{-10pt}
\end{figure*}

\subsubsection{Masking Region Selection} 
We compare four region selection methods under the same 30\% masking ratio and identical replacement method. In addition to the top-attention masking, we also conduct experiments on random masking, the least-attention masking and center masking. As shown in Figure~\ref{fig:efficiency} (c), top-attention masking achieves the highest AUROC and ACC among the compared region-selection methods. These results indicate that detector's performance gains arise from removing the visual evidence attended to by the model, rather than from arbitrary image perturbations.






\subsubsection{Effectiveness of different components in CADMP.}

As shown in Figure~\ref{fig:overview}, the proposed CADMP is built based on Cross-Modal Attention Drifts (CMAD) and Mask-based Verification (MV). Table~\ref{tab:pope_results} reports the hallucination detection performance of each component in our method on COCO. The results show that both attention drift and probability features provide informative signals for hallucination detection, while their combination yields the best overall performance. Specifically, on LLaVA-1.5-7B, the attention drift feature can achieve an AUROC of 85.7\%, while the probability signal obtains an AUROC of 72.8\%. By combining these complementary signals, CADMP can improve the AUROC to 87.4\% and achieve the best detection performance. The same trend is also observed on Qwen2.5-VL-7B and InternVL-2.5-4B. 

\begin{table}[t]
\centering
\small
\setlength{\tabcolsep}{4pt}
\begin{tabular}{lp{1.0cm}<{\centering}p{1.0cm}<{\centering}p{1.3cm}<{\centering}p{1.3cm}<{\centering}}
\toprule
\textbf{Model} & CMAD & MV & \textbf{AUC (\%) } & \textbf{ACC (\%) } \\
\midrule
\multirow{3}{*}{LLaVA-1.5-7B} & \ding{51} & & 85.7 & 80.0 \\
& & \ding{51} & 72.8 & 76.2 \\
& \ding{51} & \ding{51} & \textbf{87.4} & \textbf{82.1} \\
\midrule
\multirow{3}{*}{Qwen2.5-VL-7B}
& \ding{51} &  & 80.3 & 87.3 \\
& & \ding{51} & 66.1 & 86.8 \\
& \ding{51} & \ding{51} & \textbf{82.6} & \textbf{87.6} \\
\midrule
\multirow{3}{*}{InternVL-2.5-4B}
& \ding{51} & & 86.3 & 88.0 \\
& & \ding{51} & 57.8 & 88.0 \\
& \ding{51} & \ding{51} & \textbf{87.1} & \textbf{89.1} \\
\bottomrule
\end{tabular}
\caption{Ablation study of CMAD and MV modules for CADMP with different LVLMs.}
\label{tab:pope_results}
\vspace{-10pt}
\end{table}


\subsubsection{Ablation Study of KL Drift Formulations. } To investigate which cross-layer KL formulation is most suitable for hallucination detection, we conducted an ablation study on the Pascal VOC dataset using four drift variants over the cross-modal attention sequence $SA=\{A_{1}, A_{2}, \ldots, A_{L}\}$, where $A_i$ denotes the attention distribution at layer $i$. Specifically, \textit{Adjacent-layer} computes $KL(A_i \| A_{i-1})$ to capture local layer-to-layer changes, \textit{First-anchor} computes $KL(A_i \| A_1)$ to measure deviations from the initial grounding state, \textit{Last-anchor} computes $KL(A_i \| A_L)$ to reflect discrepancies from the final attention state, and \textit{First-last} computes $KL(A_1 \| A_L)$ as a global summary of cross-layer change. As shown in Table \ref{tab:voc_kl_ablation}, experiments on both LLaVA-1.5-7B and Qwen2.5-VL-7B show that \textit{Adjacent-layer} consistently achieves the best performance among the four variants. This suggests that hallucination is better characterized by adjacent cross-layer attention drift than by deviations from a fixed anchor or a single global discrepancy, highlighting the importance of modeling fine-grained grounding transitions.

\begin{table}[t]
  \centering
  \small
  \setlength{\tabcolsep}{3.5pt}
  \renewcommand{\arraystretch}{1.05}
  \begin{tabular}{llp{1.5cm}<{\centering}p{1.5cm}<{\centering}}
    \toprule
    \textbf{Model}
    & \textbf{Method}
    & \textbf{AUROC (\%)}
    & \textbf{ACC (\%)} \\
    \midrule

    \multirow{4}{*}{LLaVA-1.5-7B}
    & Adjacent-layer & \textbf{94.9} & \textbf{94.5} \\
    & Last-anchor    & 93.9          & 93.5          \\
    & First-anchor   & 93.2          & 93.3          \\
    & First-last     & 88.6          & 92.9          \\
    \midrule

    \multirow{4}{*}{Qwen2.5-VL-7B}
    & Adjacent-layer & \textbf{92.1} & \textbf{96.5} \\
    & Last-anchor    & 90.4          & 94.6          \\
    & First-anchor   & 86.3          & 93.9          \\
    & First-last     & 84.1          & 90.1          \\
    \bottomrule
  \end{tabular}
  \caption{Ablation study of different KL drift formulations.}
  \label{tab:voc_kl_ablation}
  \vspace{-10pt}
\end{table}

\subsubsection{Visualization Analysis.}
We present qualitative visualizations in Figure~\ref{fig:case_analysis}. As shown, hallucinated cases exhibit substantially stronger cross-modal attention drifts in the critical intermediate layers and smaller probability drops after masking relevant visual regions, indicating weaker visual grounding. In contrast, grounded cases show relatively milder attention changes and greater sensitivity to visual perturbation. These results further support the effectiveness of our design in distinguishing hallucinated and grounded cases.

\section{Conclusion}

In this paper, we proposed CADMP, a hallucination detection framework for large vision-language models. CADMP detects object hallucinations by measuring distribution shifts between layers and further verifying the visual grounding of model predictions through mask-based probability variation analysis. By jointly capturing internal attention drifts and perturbation-based visual dependence, our method provides a more fine-grained and complementary empirical signal for hallucination detection. Extensive experiments on three representative LVLMs, including LLaVA-1.5-7B, Qwen2.5-VL-7B, and InternVL-2.5-4B, demonstrate the effectiveness and generalizability of the proposed method across different hallucination detection settings.

\bibliography{main}

\clearpage
\appendix
\section{Further Ablation Studies}

\subsection{Effect of Masking Ratios}
We investigate the effect of the masking ratio $\tau$ on the MSCOCO dataset using LLaVA-1.5-7B by varying $\tau$ from 10\% to 50\%. As shown in Table~\ref{tab:masking_ratios}, CADMP achieves the best performance at a masking ratio of 30\%, obtaining an AUROC of 87.97\% and an ACC of 82.88\%. These results suggest that a small masking ratio may fail to cover the key object regions, whereas a large ratio may introduce substantial visual perturbations. A masking ratio of 30\% achieves a favorable balance between removing critical visual evidence and preserving the overall image context.

\begin{table}[htbp]
    \centering
    \begin{tabular}{cccc}
        \toprule
        Masking Ratio $\tau$ & AUROC (\%) & ACC (\%) \\
        \midrule
        10\% & 87.17 & 82.32  \\
        20\% & 87.27 & 81.84 \\
        30\% & \textbf{87.97} & \textbf{82.88}  \\
        40\% & 87.26 & 82.44  \\
        50\% & 87.31 & 81.57  \\
        \bottomrule
    \end{tabular}
    \caption{Effect of different masking ratios on the MSCOCO dataset using LLaVA-1.5-7B.}
    \label{tab:masking_ratios}
\end{table}

\subsection{Effect of Mask Filling Values}

To study the choice of fill color for masked image regions, we conduct an experiment using LLaVA-1.5-7B on the MSCOCO dataset. As shown in Table~\ref{tab:mask_filling_values}, filling the masked regions with the default gray value of 128 achieves the best performance, with an AUROC of 87.97\% and an ACC of 82.88\%. Compared with black, white and the ImageNet mean RGB value $(123,116,103)$, gray filling achieves the best performance. Therefore, we adopt gray filling with a value of 128, as it effectively masks the selected visual evidence without introducing excessively dark or bright perturbations.

\begin{table}[htbp]
    \centering
    \small
    \begin{tabular}{lccc}
        \toprule
        Filling Value  & AUROC (\%) & ACC (\%) \\
        \midrule
        Gray  & \textbf{87.97} & \textbf{82.88} \\
        Black             & 87.23 & 82.02 \\
        White           & 87.38 & 82.02 \\
        Mean         & 87.31 & 81.72 \\
        \bottomrule
    \end{tabular}
    \caption{Effect of different mask filling values on the MSCOCO dataset using LLaVA-1.5-7B.}
    \label{tab:mask_filling_values}
\end{table}

\subsection{Multi-token Objects}
To compute the KL divergence sequence and the probability variation for a  target object token, we select the first token of each multi-token object. To evaluate this design choice, we conduct an ablation study comparing three strategies: using the first token, using the last token and taking the average across all tokens. We conduct experiments on MSCOCO dataset across three LVLMs in Table~\ref{tab:token_selection}. The results show that the first-token  strategy is most effective, likely because the first token often captures the core semantic meaning of the object.

\begin{table}[htbp]
    \centering
    \small
    \setlength{\tabcolsep}{2pt}
    \begin{tabular}{lccc}
        \toprule
        Strategy 
        & LLaVA-1.5-7B 
        & Qwen2.5-VL-7B 
        & InternVL2.5-4B \\
        \midrule
        First   
        & \textbf{87.97} 
        & \textbf{82.60} 
        & \textbf{88.47} \\
        
        Last    
        & 86.34 
        & 81.46 
        & 88.17 \\
        
        Average 
        & 87.65 
        & 81.97 
        & 88.16 \\
        \bottomrule
    \end{tabular}
    \caption{AUROC (\%) of different token selection strategies on the MSCOCO dataset.}
    \label{tab:token_selection}
\end{table}

\subsection{Ablation of Drift Feature Representations}

To examine whether the complete layer-wise drift sequence is necessary for hallucination detection, we compare six KL feature representations. Let $K_i=D_{\mathrm{KL}}(A_i\|A_{i-1})$ denote the attention drift between two adjacent decoder layers, and let
$k^{*}=\arg\max_i K_i$ denote the position of the largest drift. The Full KL sequence uses the complete sequence $[K_2,\ldots,K_L]$. Maximum KL and Mean KL compress the sequence into its maximum and average values, respectively, while Statistical summary represents it using the mean, maximum, and standard deviation. For the Peak-centered local window variants, we select three or five consecutive KL values centered around $k^{*}$, capturing the local drift pattern around the strongest transition.

\begin{table}[t]
\centering
\small
\setlength{\tabcolsep}{4pt}
\begin{tabular}{lcc}
\toprule
\textbf{Variant} 
& \textbf{AUROC (\%)} 
& \textbf{ACC (\%)} \\
\midrule
\textbf{Full KL sequence}
& $\mathbf{87.97 }$
& $\mathbf{82.88 }$ \\

Statistical summary
& $80.58 $
& $79.01 $ \\

Peak centered (5)
& $79.05$
& $78.23 $ \\

Mean KL
& $78.20$
& $78.62 $ \\

Peak centered (3)
& $76.50 $
& $77.94 $ \\

Maximum KL
& $72.68 $
& $76.27$ \\
\bottomrule
\end{tabular}
\caption{Ablation of KL-drift feature representations on MSCOCO using LLaVA-1.5-7B.}
\label{tab:drift_representation}
\end{table}

As shown in Table~\ref{tab:drift_representation}, the full KL sequence achieves the best performance, reaching an AUROC of $87.97\%$ and an ACC of $82.88\%$. Compressing the sequence into statistical summaries decreases the AUROC by $7.39$ percentage points, while using only the maximum KL value leads to a substantial reduction of $15.29$ points. These results indicate that hallucination information is distributed across the layer-wise drift sequence rather than being confined to the largest drift or its immediate neighboring layers.

\subsection{Distribution of Peak Attention-Drift Layers}

\begin{figure}[t]
    \centering
    \includegraphics[width=\linewidth]{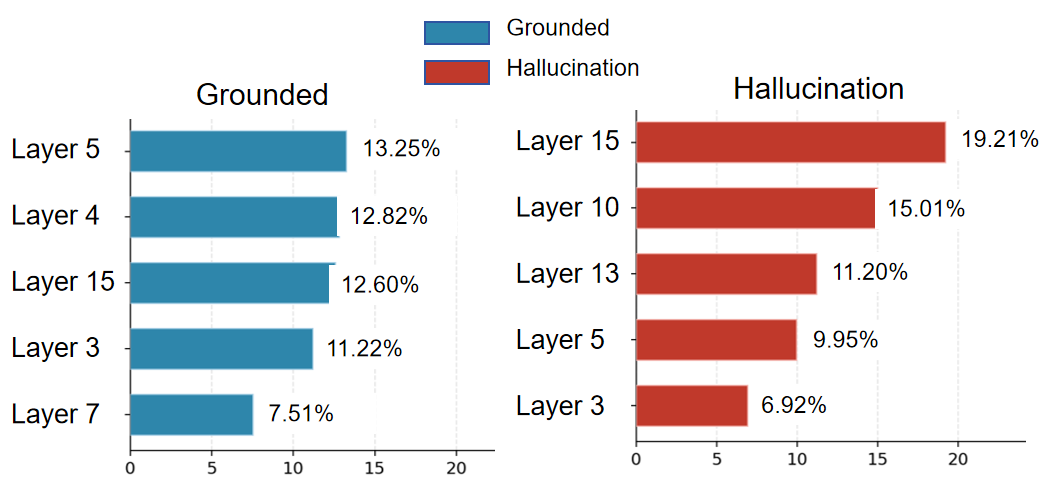}
    \caption{Distribution of the layers exhibiting the maximum adjacent-layer cross-modal attention drift for grounded and hallucinated samples on Qwen2.5-VL-7B.}
    \label{fig:peak_layer_distribution}
\end{figure}

Figure~\ref{fig:peak_layer_distribution} shows the distribution of layers where the maximum adjacent-layer attention drift occurs on Qwen2.5-VL-7B. For hallucinated samples, the maximum attention drift most frequently occurs at Layers 15, 10, and 13, suggesting that abrupt cross-modal attention redistribution tends to emerge in the intermediate layers. This observation further supports the effectiveness of CADMP in capturing hallucination-related signals from layer-wise attention dynamics.

\section{Experiment Datasets}

\subsection{COCO-Caption}
To evaluate hallucination detection in open-ended generation, we construct the COCO-Caption setting from the COCO-val2014 split of MSCOCO~\cite{lin2014microsoft}. We select 5,000 images from COCO-val2014 and independently generate captions using each evaluated LVLM. Caption generation is performed with greedy decoding, and the maximum number of newly generated tokens is set to 128. We apply the CHAIR annotation protocol~\cite{rohrbach2018object} to extract object mentions from each generated caption and determine whether each mentioned object is supported by the corresponding image. For multi-token objects, CADMP uses only the first subtoken for probability, attention, and masking.

Because one generated caption may contain multiple object mentions, directly splitting object-level instances could place objects extracted from the same image in different subsets and introduce information leakage. We therefore perform an image-level split using a ratio of 70\%/10\%/20\% for training, validation, and testing, respectively. 

\subsection{POPE}
POPE~\cite{li2023evaluating} is designed to evaluate object-existence hallucinations in Large Vision-Language Models. It formulates the evaluation as a binary question-answering task by asking whether a particular object is present in an input image. The benchmark is constructed from images in MSCOCO~\cite{lin2014microsoft} and contains three subsets, namely Random, Popular, and Adversarial. The Random subset uniformly samples objects that do not appear in the image, the Popular subset selects frequently occurring object categories, and the Adversarial subset selects absent objects that commonly co-occur with the objects present in the image.

For implementation, we use 1,500 images. For each image, three positive questions and three negative questions are constructed, where positive questions query objects present in the image and negative questions query absent objects. This results in a total of 9,000 image-question pairs. 

\subsection{Pascal VOC}

Pascal VOC is a widely used benchmark for object detection and recognition in computer vision~\cite{everingham2010pascal}. It provides image-level and object-level annotations for 20 common object categories, covering animals, vehicles, household objects, and people. Owing to its diverse visual scenes and reliable category annotations, Pascal VOC offers a suitable testbed for evaluating whether an LVLM can correctly determine the presence or absence of specific objects.

In our experiments, we use images from the VOC2012 split to construct an additional evaluation set for object hallucination detection. Specifically, for each image, we create one positive sample by querying an object category that is present according to the ground-truth annotations, together with three negative samples based on categories absent from the image. This results in a total of 9,000 samples in the Pascal VOC-based evaluation set. 

\subsection{AMBER}

AMBER~\cite{wang2023amber} is a benchmark designed to evaluate hallucinations in multimodal large language models without relying on an external language model as the evaluator. It contains 1,004 images and provides both generative and discriminative evaluation settings. While the generative setting assesses hallucinations in free-form model responses, the discriminative setting uses targeted questions to examine whether a model can accurately distinguish visually supported information from unsupported claims.

The discriminative component covers three representative hallucination types: object existence, attribute, and relation hallucinations. Object-existence hallucination occurs when a model claims that an object is present even though it does not appear in the image. Attribute hallucination refers to assigning an incorrect property, such as color, number, size, or state, to an existing object. Relation hallucination occurs when the model incorrectly describes the spatial or semantic relationship between objects. By jointly evaluating these three dimensions, AMBER provides a broader and more fine-grained assessment than benchmarks focusing only on object presence.

\section{Baselines}

We compare CADMP with six representative object hallucination detection baselines.

\subsection{Negative Log-Likelihood (NLL)}
NLL~\cite{zhou2023analyzing} uses the generation probability of the target object token as the hallucination score, where $j$ denotes the positional index of object $o$:
\[
s_{\mathrm{NLL}}(o)
=
-\log P(o \mid \mathbf{v},\mathbf{t},Y_{<j}).
\]
For a multi-token object, we use its first token. A larger NLL score indicates a higher likelihood of hallucination.

\subsection{Entropy}
Entropy~\cite{malinin2020uncertainty} measures predictive uncertainty from the vocabulary distribution at the target object-token position:
\[
s_{\mathrm{Ent}}(o)
=
-\sum_{w\in\mathcal{V}} P(w)\log P(w),
\]
where $\mathcal{V}$ denotes the vocabulary. Higher entropy indicates greater uncertainty in the object prediction.

\subsection{IC}
IC~\cite{jiang2025interpreting} applies the Visual Logit Lens to project the hidden representation of each visual token at every decoder layer into the vocabulary space:
\[
\mathrm{VLL}_{l}(v_i)
=
\mathbf{h}_{l}(v_i)\mathbf{W}_{U},
\]
where $\mathbf{h}_{l}(v_i)$ denotes the hidden representation of visual token $v_i$ at layer $l$, and $\mathbf{W}_{U}$ is the language-model unembedding matrix. The IC score is defined as the maximum probability assigned to the target object token across all decoder layers and visual positions:
\[
s_{\mathrm{IC}}(o)
=
\max_{l,i}
\mathrm{Softmax}
\left(
\mathrm{VLL}_{l}(v_i)
\right)_{o}.
\]
A higher IC score indicates stronger internal visual support for the presence of the target object.

\subsection{SVAR}
SVAR~\cite{jiang2025devils} measures the visual attention received by a generated object token. For an object token $o$, the Visual Attention Ratio at layer $\ell$ and attention head $h$ is defined as
\[
\mathrm{VAR}_{\ell,h}(o)
=
\sum_{i=1}^{N} A_{\ell,h}(o,v_i).
\]
where $A^{(\ell,h)}(o,v_i)$ denotes the attention weight from the object token $o$ to the $i$-th visual token $v_i$, and $N$ is the number of visual tokens. SVAR then averages the VAR scores across all attention heads and sums them over Layers 5--18:
\[
s_{\mathrm{SVAR}}(o)
=
\frac{1}{H}
\sum_{\ell=5}^{18}
\sum_{h=1}^{H}
\operatorname{VAR}^{(\ell,h)}(o),
\]
where $H$ denotes the number of attention heads. A higher SVAR score indicates that the generated object token assigns more attention to visual information.

\subsection{GLSim}
GLSim~\cite{park2025glsim} is an object hallucination detection method that jointly evaluates global scene-level consistency and local visual grounding. It measures whether a predicted object is semantically compatible with the overall image context and whether relevant visual regions provide sufficient evidence for its presence. For evaluation on POPE, we extract the queried object from each question and use its token representation to compute the GLSim score.

\subsection{DHCP}
DHCP~\cite{zhang2025dhcp} detects object hallucinations by modeling cross-modal attention patterns within LVLMs. It uses the attention assigned by generated tokens to visual tokens across decoder layers as internal evidence to distinguish visually grounded predictions from hallucinated ones.

\subsection{Baselines Details}

For experiments on LLaVA-1.5-7B using coco-caption dataset, we implement all baselines using the following configurations.  For SVAR, we aggregate the attention assigned by the generated object token to all visual tokens over Layers 5-18 and normalize the resulting score by the number of attention heads. For GLSim, we set the image layer and text layer to 32 and 31, respectively, select the top 32 visual tokens for local similarity computation, and combine the global and local similarity scores using weights of 0.6 and 0.4; the visual sequence contains 576 tokens arranged in a $24 \times 24$ grid. DHCP extracts  head-averaged attention from the generated object token to the visual tokens at each layer, pads each layer to 576 visual positions, and flattens the resulting features. Its MLP classifier consists of a 128-dimensional hidden layer followed by a two-class output layer and is trained using Adam with a learning rate of $1 \times 10^{-3}$, a batch size of 256, and 30 epochs. We use cross-entropy loss and a Weighted Random Sampler with inverse class-frequency weights.

\section{Experiment Settings}

\subsection{Models}
We evaluate CADMP on three representative open-source LVLMs, including LLaVA-1.5-7B, Qwen2.5-VL-7B-Instruct, and InternVL2.5-4B, which differ in model architecture, visual-language alignment strategy, and parameter scale. All experiments are conducted using Python 3.10 and PyTorch 2.6.0 on a single NVIDIA A6000 GPU with 48GB of memory.

\subsection{Mask Construction}

After identifying the layer transition with the maximum attention drift, we select its preceding layer as the suspicious layer. We extract the cross-modal attention from the target output token to all visual tokens and reshape it into an $H \times W$ attention map according to the spatial layout of the visual feature map. The resulting attention map is then upsampled to the original image resolution using bilinear interpolation, such that each pixel is assigned a continuous attention weight. Next, a threshold is determined according to a predefined percentile of the upsampled attention values. Pixels whose attention weights are greater than or equal to this threshold are replaced with a neutral gray value of $128$, while the remaining pixels are left unchanged. The resulting masked image is subsequently fed into the LVLM for a second forward pass to measure the change in the prediction probability of the target token.

\subsection{Implementation Details}

For the COCO-Caption experiments, LLaVA-1.5-7B, Qwen2.5-VL-7B, and InternVL2.5-4B use the same MLP detector configuration. The input consists of the layer-wise KL-divergence drift sequence and three probability-based features: the target-token probability under the original image, the probability under the masked image, and their difference. All features are standardized using StandardScaler before being passed to a two-hidden-layer MLP. The two hidden layers contain 128 and 64 neurons, respectively, and each is followed by Batch Normalization, ReLU activation, and Dropout with a rate of 0.3. The detector is trained using AdamW with a learning rate of $5 \times 10^{-4}$, a batch size of 128, and a weight decay of $1 \times 10^{-4}$. Training is performed for at most 150 epochs, with early stopping using a patience of 20 epochs to select the best checkpoint on the validation set. The data are divided into training, validation, and test sets using a ratio of 70\%/10\%/20\%.

For AUROC computation, hallucinated samples are treated as the
positive class. For threshold-based evaluation, predictions are
obtained from the probability of the grounded class, with the
decision threshold selected on the validation set by maximizing
the F1 score. All experiments are conducted using three random seeds: 42, 7777, and 37, and the final results are reported as the mean and standard deviation across the three runs.

\section{Visualization of CADMP}

\begin{figure*}[t!]
  \centering
  \includegraphics[width=\linewidth]{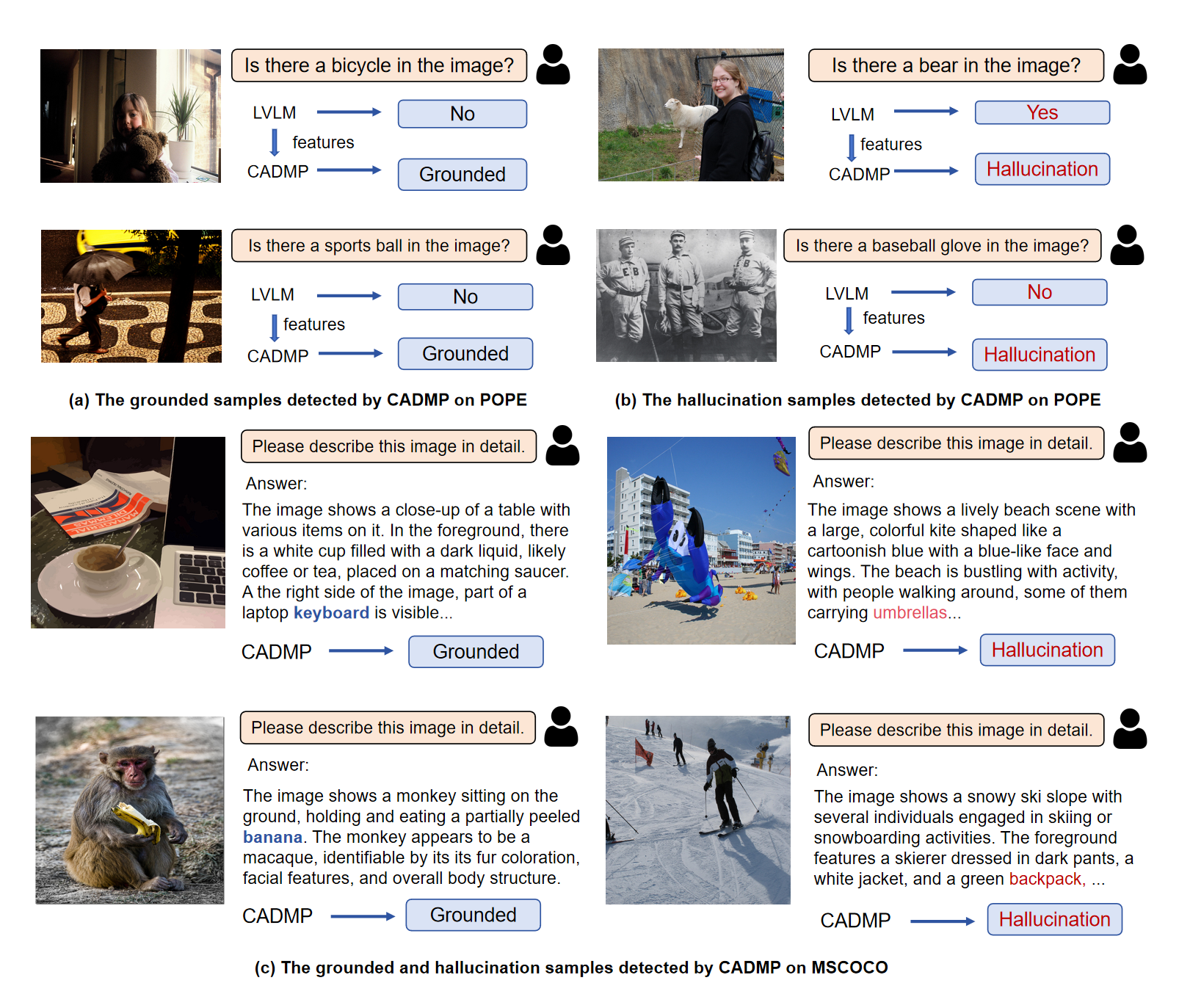}
  \caption{Visualization of CADMP for detecting hallucinations on the POPE and COCO-Caption datasets} 
  \label{fig:visual}
\end{figure*}

To qualitatively examine the effectiveness of CADMP, we present visualization examples from both the POPE and COCO-Caption datasets in Figure~\ref{fig:visual}. 

\end{document}